\documentclass[11pt]{article}
\usepackage{acl2014}
\usepackage{times}
\usepackage{url}
\usepackage{latexsym}
\usepackage{xspace}
\usepackage{booktabs}
\usepackage{color}
\usepackage{graphicx}
\usepackage{tabulary}
\usepackage{enumitem}

\usepackage{amsfonts}

  {\begin{itemize}[topsep=0pt, partopsep=0pt] %
    \setlength{\itemsep}{0pt}%
    \setlength{\parskip}{0pt}%
    }%
  {\end{itemize}}
  
  \usepackage{color}
  {\begin{enumerate}≈ \footnotesize%
    \setlength{\itemsep}{0pt}%
    \setlength{\parskip}{0pt}%
    }%
  {\end{enumerate}}

\usepackage{microtype}
\usepackage{multirow}
\usepackage{verbatim}
\usepackage{amsmath,amsthm,amssymb}
\usepackage{array}
\usepackage[scaled=0.86]{helvet}
\usepackage{ifthen}
\usepackage{courier}
\usepackage[linesnumbered,vlined,ruled]{algorithm2e}

\newcommand{\captionfonts}{\small}
\makeatletter  
\long\def\@makecaption#1#2{%
  \vskip\abovecaptionskip
  \sbox\@tempboxa{{\captionfonts #1: #2}}%
  \ifdim \wd\@tempboxa >\hsize
    {\captionfonts #1: #2\par}
  \else
    \hbox to\hsize{\hfil\box\@tempboxa\hfil}%
  \fi
  \vskip\belowcaptionskip}
\makeatother   

\belowdisplayskip \abovedisplayskip

\newcommand{\sts}{{{\textsc{Seq2Seq}}}\xspace}

\title{Visualizing and Understanding Neural Models in NLP}

\author{Jiwei Li$^1$, Xinlei Chen$^2$, Eduard Hovy$^2$ and Dan Jurafsky$^1$\\
{\normalsize$^1$Computer Science Department, Stanford University, Stanford, CA 94305, USA}\\
{\normalsize$^2$Language Technology Institute, Carnegie Mellon University, Pittsburgh, PA 15213, USA}\\
{\normalsize \{jiweil,jurafsky\}@stanford.edu~~~~~~\{xinleic,ehovy\}@andrew.cmu.edu}
}

\begin{document}
\maketitle
\begin{abstract}
%
%
%
While neural networks have been successfully applied to many NLP tasks
the resulting vector-based models are very difficult to interpret.
For example it's not clear how they achieve {\em compositionality},
building sentence meaning from the meanings of words and phrases.
In this paper we describe strategies for visualizing compositionality 
in neural models for NLP, inspired by similar work in computer vision.
We first plot unit values to visualize compositionality of negation, intensification, and 
concessive clauses, allowing us to see well-known markedness asymmetries in negation.
We then introduce methods for visualizing a unit's {\em salience},
the amount that
it contributes to the final composed meaning from first-order derivatives. 
Our general-purpose methods may have wide applications for understanding
compositionality and other semantic properties of deep networks.
\end{abstract}

\section{Introduction} 

Neural models match or outperform the performance of 
other state-of-the-art systems on a variety of NLP tasks. 
Yet unlike traditional feature-based classifiers 
that assign and optimize weights to varieties of human interpretable features 
(parts-of-speech, named entities, word shapes, syntactic parse features etc)
the behavior of deep learning models is much less easily interpreted.
Deep learning models mainly operate on word
embeddings (low-dimensional, continuous, real-valued vectors) through
multi-layer neural architectures, each layer of which is  characterized
as an array of hidden neuron units.
It is unclear how deep learning models 
deal with {\em composition}, implementing functions like
negation or intensification, or combining
meaning from different parts of the sentence,
filtering away the informational chaff from the wheat, 
to build sentence meaning.

In this paper, we explore 
multiple strategies  to interpret meaning
composition in neural models.
We employ traditional methods like representation plotting,
and  introduce simple  strategies for measuring 
how much a neural unit contributes to meaning composition, 
its `salience' or importance 
using first derivatives.

Visualization techniques/models represented in this work shed important  light on how neural models work:
For example, we illustrate that
LSTM's success is due to its ability in 
maintaining a much sharper focus on the important key words than other models; 
Composition in multiple clauses works competitively, and
that the models are able to capture negative asymmetry, an important property
of semantic compositionally in natural language understanding; 
there is sharp dimensional locality, with certain dimensions marking
negation and quantification in a manner that was surprisingly localist. 
Though our attempts only touch superficial points in neural models,
and each method has its pros and cons, 
together they may offer
some insights into the behaviors of neural models in language based tasks,
marking one initial step toward understanding how they achieve
meaning composition in natural language processing.

The next section describes some visualization models in vision and NLP that have inspired this work.
We describe datasets and the adopted neural models in Section 3. 
Different visualization strategies and correspondent analytical results are presented separately in Section 4,5,6, followed by a brief conclusion. 
\section{A Brief Review of Neural Visualization}

Similarity is commonly visualized graphically, generally
by projecting the embedding
space into two dimensions and observing that similar words tend to be clustered
together (e.g., \newcite{elman89}, \newcite{ji2014representation}, \newcite{faruqui2014improving}).
\cite{karpathy2015visualizing} attempts to interpret recurrent neural models from a statical point of view and does deeply touch compositionally of meanings. 
Other relevant attempts include \cite{fyshe2015compositional,faruqui2015sparse}.

Methods for interpreting and visualizing neural models have been much more
significantly explored in vision, especially for
Convolutional Neural Networks (CNNs or ConvNets) \cite{krizhevsky2012imagenet},
multi-layer neural networks in which the original matrix of image
pixels is convolved and pooled as it is passed on to hidden layers.
ConvNet visualizing techniques consist mainly in mapping
the different layers of the network
(or other features like SIFT \cite{lowe2004distinctive} and HOG
\cite{dalal2005histograms}) back to the initial image input,
thus capturing the human-interpretable information they represent in the input, 
and how units in these layers contribute to any final decisions
\cite{simonyan2013deep,mahendran2014understanding,nguyen2014deep,szegedy2013intriguing,girshick2014rich,zeiler2014visualizing}. Such methods include:

(1) Inversion: Inverting the representations by training an additional model 
to project outputs from different neural levels 
back to  the initial input images
 \cite{mahendran2014understanding,vondrick2013hoggles,weinzaepfel2011reconstructing}. The 
intuition behind reconstruction is that 
the pixels that are reconstructable from the current representations
are the content of the representation. The inverting algorithms allow
the current representation  to align with  corresponding parts of the original images.

(2) Back-propagation \cite{erhan2009visualizing,simonyan2013deep} and Deconvolutional Networks \cite{zeiler2014visualizing}:  Errors are back propagated from output layers to each intermediate
layer and finally to the original image inputs. 
Deconvolutional Networks work in a similar way by projecting outputs
back to initial inputs layer by layer, each layer associated
with one supervised model for projecting upper ones to lower ones
These strategies make it possible to spot 
active regions or ones that contribute the most to the final classification decision.

(3) Generation:  This group of work generates images in a specific class
from a sketch guided by already trained neural models \cite{szegedy2013intriguing,nguyen2014deep}.
Models begin with an image whose pixels are randomly initialized
and mutated  at each step.  The specific layers that are activated
at different stages of image construction  can help in interpretation.

While the above strategies inspire the work we present in this paper,
there are fundamental differences between vision and NLP.
In NLP words function as basic units, and hence (word) vectors rather than single
pixels are the basic units.
Sequences of words (e.g., phrases and sentences) are also presented in a more 
structured way than arrangements of pixels. 
In parallel to our research, independent researches \cite{karpathy2015visualizing} have been conducted to 
explore similar direction from an error-analysis point of view, by analyzing 
 predictions and errors from a recurrent neural models. 
 Other distantly relevant works include: 
\newcite{murphy2012learning,fyshe2015compositional}
used an manual task to quantify the interpretability
of semantic dimensions by presetting human users with a list of words and ask them to choose the one that does not belong to the list. 
\newcite{faruqui2015sparse}. Similar strategy is adopted in \cite{faruqui2015sparse} by extracting top-ranked words in each vector dimension.

\section{Datasets and Neural Models}
We explored two datasets on which neural models are trained, one of which is of relatively small scale and the other of large scale. 
\subsection{Stanford Sentiment Treebank} 
Stanford Sentiment Treebank
is
a benchmark dataset widely used for neural model evaluations.
The dataset contains gold-standard sentiment labels for every parse tree constituent, from  sentences to phrases to individual words,  for 215,154 phrases in 11,855 sentences.
The task is to perform both fine-grained (very positive, positive, neutral, negative and very negative)
and coarse-grained (positive vs negative) classification at both the phrase and sentence level. 
For more details about the dataset, please refer to \newcite{socher2013recursive}.
 
While many studies on this dataset use recursive parse-tree models,
in this work we employ only standard sequence models (RNNs and LSTMs) since these are the most
widely used current neural models, and sequential visualization is more straightforward.
We therefore first transform each parse tree node to a sequence of tokens. The sequence is
first mapped to a phrase/sentence representation 
and fed into a softmax classifier.  Phrase/sentence representations are built
with the following three models:
{\it Standard Recurrent Sequence} with \textsc{tanh} activation functions,
{\it LSTMs} and {\it Bidirectional LSTMs}. 
For details about the three models, please refer to Appendix. 

\paragraph{Training}
AdaGrad with mini-batch was used for training, with parameters 
($L2$ penalty, learning rate, mini batch size) tuned on the development set.
The number of iterations is treated as a variable to tune and
parameters are harvested based on the best performance on the
dev set. The number of dimensions for the word and hidden layer are set
to 60 with 0.1 dropout rate. Parameters are tuned on the dev set.
The standard recurrent model achieves 0.429 (fine grained) and 0.850 (coarse grained)
accuracy at the sentence level;  LSTM achieves
0.469 and 0.870, and Bidirectional LSTM 0.488 and 0.878, respectively.

\subsection{Sequence-to-Sequence Models}
\sts
are neural models aiming at generating a sequence of output texts given inputs. 
Theoretically, \sts models can be adapted to NLP tasks that can be formalized as predicting outputs given inputs 
and serve for different purposes due to different inputs and outputs, e.g., machine translation where inputs correspond to source sentences and outputs to target sentences \cite{sutskever2014sequence,luong2014addressing}; 
conversational response generation if inputs correspond to messages and outputs correspond to responses \cite{vinyals2015neural,li2015diversity}. 
\sts need to be trained on massive amount of data 
for 
 implicitly  semantic and
syntactic relations between pairs to be learned.

\sts models map an input sequence to a vector representation using LSTM models and then 
sequentially predicts tokens based on  the pre-obtained representation. 
The model defines a distribution over outputs (Y) and sequentially predicts tokens
given inputs (X)
 using a softmax function.
\begin{equation*}
\begin{aligned}
P(Y|X)
&=\prod_{t=1}^{n_y}p(y_t|x_1,x_2,...,x_t,y_1,y_2,...,y_{t-1})\\
&=\prod_{t=1}^{n_y}\frac{\exp(f(h_{t-1},e_{y_t}))}{\sum_{y'}\exp(f(h_{t-1},e_{y'}))}
\end{aligned}
\label{equ-lstm}
\end{equation*}
where $f(h_{t-1}, e_{y_t})$ denotes the activation function between $h_{t-1}$ and $e_{y_t}$, where $h_{t-1}$ is the representation output from the LSTM at time \mbox{$t-1$}. 
For each time step in word prediction, \textsc{Seq2Seq} models combine the current token with previously
built embeddings for next-step word prediction.

For easy visualization purposes, we turn to the 
most straightforward task---autoencoder---
 where inputs and outputs are identical. The goal of an autoencoder is to reconstruct inputs from the pre-obtained representation. 
 We would like to see how individual input tokens affect the overall sentence representation and each of the tokens to predict in outputs. 
We trained the auto-encoder on a subset of WMT'14 corpus containing 4 million english sentences with an average length of 22.5 words. 
We followed training protocols described in \cite{sutskever2014sequence}.

\section{Representation Plotting }
\begin{figure}[htb]
\centering
{\bf Intensification}
\includegraphics[width=3in]{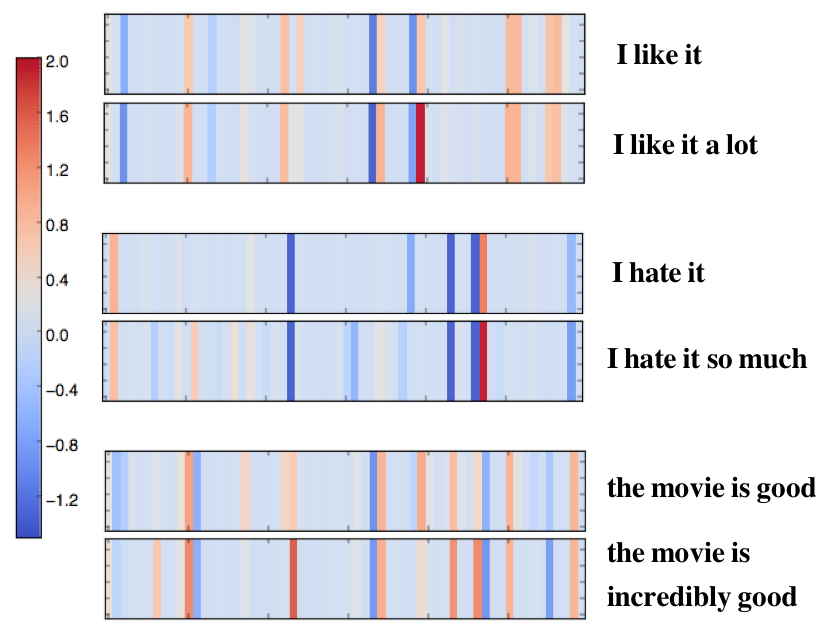}
{\bf Negation}
\includegraphics[width=3in]{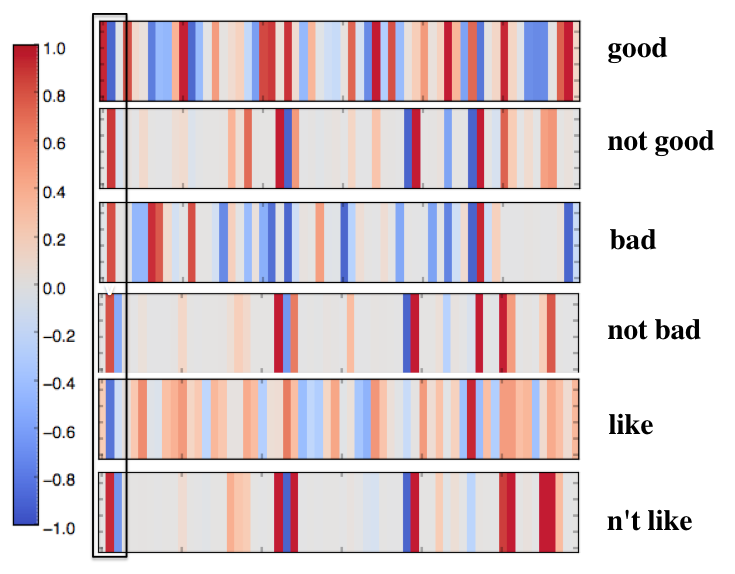}
\caption{Visualizing intensification and negation. Each vertical bar shows
the value of one dimension in the final sentence/phrase representation after compositions. Embeddings for phrases or sentences are attained by composing word representations from the pretrained model. }
\label{negation}
\end{figure}

We begin with simple plots of representations to shed light on local compositions using Stanford Sentiment Treebank.
\paragraph{Local Composition}
Figure \ref{negation}
shows a 60d heat-map vector for the representation of selected words/phrases/sentences,
with an emphasis on
extent modifications (adverbial and adjectival) 
and negation. Embeddings for phrases or sentences are attained by composing word representations from the pretrained model. 

The intensification part of Figure \ref{negation} shows  suggestive
patterns where values for a few dimensions 
are strengthened by modifiers like ``a lot" (the red bar in the first example)
``so much" (the red bar in the second example),
and ``incredibly".
Though the patterns for negations are not as clear,
there is still a consistent reversal for some dimensions, visible as a shift between blue
and red for dimensions boxed on the left.

\begin{figure*}[!htb]
\centering
\includegraphics[width=5in]{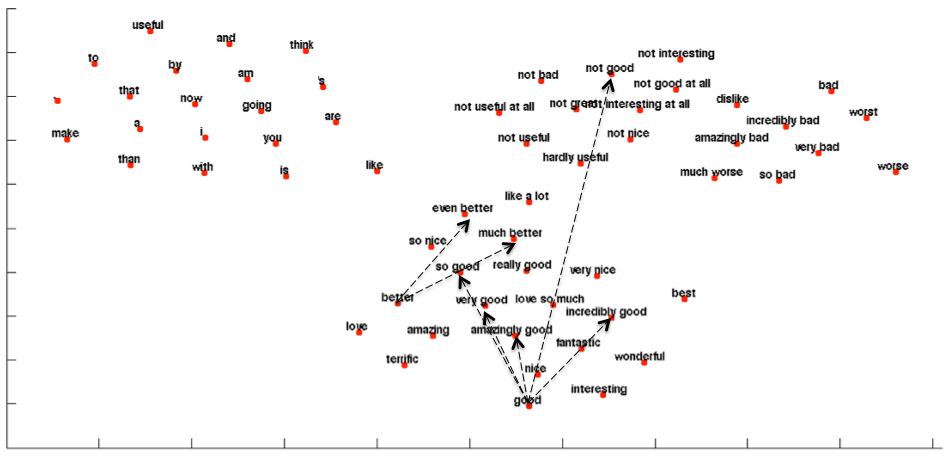}
\caption{t-SNE Visualization on latent representations for modifications and negations.}
\label{tsne}
\end{figure*}
\begin{figure}[!htb]
\centering
\includegraphics[width=2.5in]{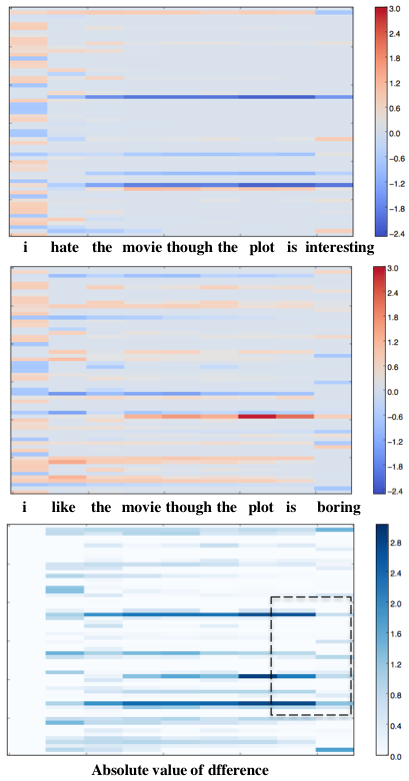}
\caption{Representations over time from LSTMs.
Each column corresponds to outputs from LSTM at each time-step (representations obtained after combining current word embedding with previous build embeddings).
Each grid from the column corresponds to each dimension of current time-step representation. 
The last rows correspond to absolute differences for each time step between two sequences. }\label{adverse}
\end{figure}

 \begin{figure*}[!ht]
\centering
\includegraphics[width=4.5in]{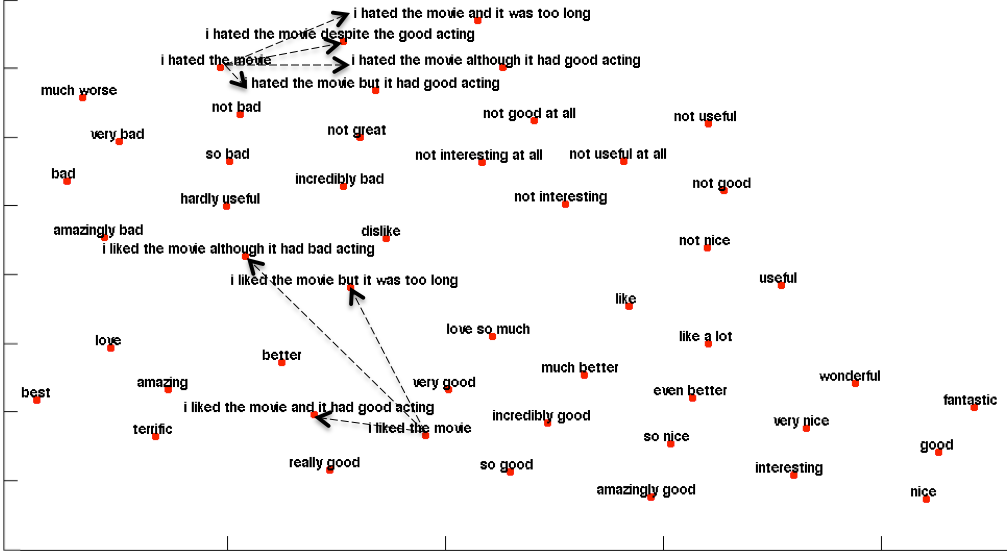}
\caption{t-SNE Visualization for clause composition.}
\label{clause}
\end{figure*}

We then visualize words and phrases using t-sne \cite{van2008visualizing}
in Figure \ref{tsne}, deliberately adding in some random words for
comparative purposes. As can be seen, neural models nicely learn
the properties of local compositionally, clustering negation+positive
words (`not nice', 'not good') together with negative words.
Note also the asymmetry of negation: ``not bad" is clustered
more with the negative than the positive words (as shown both
in Figure \ref{negation} and \ref{tsne}).
This asymmetry has been widely discussed in linguistics,
for example as arising from markedness, since `good' is the unmarked
direction of the scale \cite{clarkclark77,horn89,fraenkel08}.
This suggests that although the model does seem to focus on certain
units for negation in Figure \ref{negation},
the neural model is not just learning to apply a fixed transform for `not'
but is able to capture the subtle differences in
the composition of different words.

\paragraph{Concessive Sentences}

In concessive sentences, two clauses have opposite polarities,
usually related by a contrary-to-expectation implicature.
We plot evolving representations over time for two concessives in Figure \ref{adverse}. 
The plots  suggest:

1. For tasks like sentiment analysis whose goal is to predict a specific
semantic dimension (as opposed to general tasks like language model word prediction), 
too large a dimensionality leads to many dimensions non-functional (with values close to 0),
causing two sentences of opposite sentiment to differ only in
a few dimensions.  This may explain why more dimensions don't
necessarily lead to better performance on such tasks (For example, as reported in \cite{socher2013recursive}, 
optimal performance
 is achieved when word dimensionality is set to between 25 and 35).

2.  Both sentences contain two clauses connected by the conjunction ``though". 
Such two-clause sentences might either work
{\em collaboratively}--- models would remember the word ``though" and make
the second clause share the same sentiment orientation as first---or {\em competitively},
with the stronger one dominating. The 
region within dotted line in Figure
\ref{adverse}(a) favors the second assumption: the difference between the two
sentences is diluted when the final words (``interesting" and ``boring") appear.

\paragraph{Clause Composition}

In Figure \ref{clause} we explore this clause composition in more detail.
Representations move closer to the negative sentiment region by adding
negative clauses like
``although it had bad acting" or ``but it is too long" to the end
of a simply positive ``I like the movie".  
By contrast, adding a concessive clause to a negative clause
does not move toward the positive; ``I hate X but ..." is still very negative,
not that different than ``I hate X".
This difference again suggests the model is able to capture  negative
asymmetry \cite{clarkclark77,horn89,fraenkel08}.

\begin{figure*}[!ht]
\centering
\includegraphics[width=6.5in]{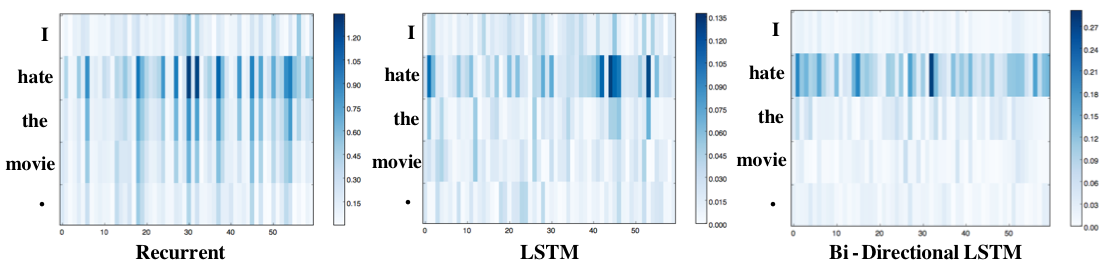}
\caption{Saliency heatmap for for ``I hate the movie ." Each row corresponds to saliency scores for the correspondent word representation with each grid representing each dimension. }\label{V1}
\end{figure*}

\begin{figure*}[!ht]
\centering
\includegraphics[width=6in]{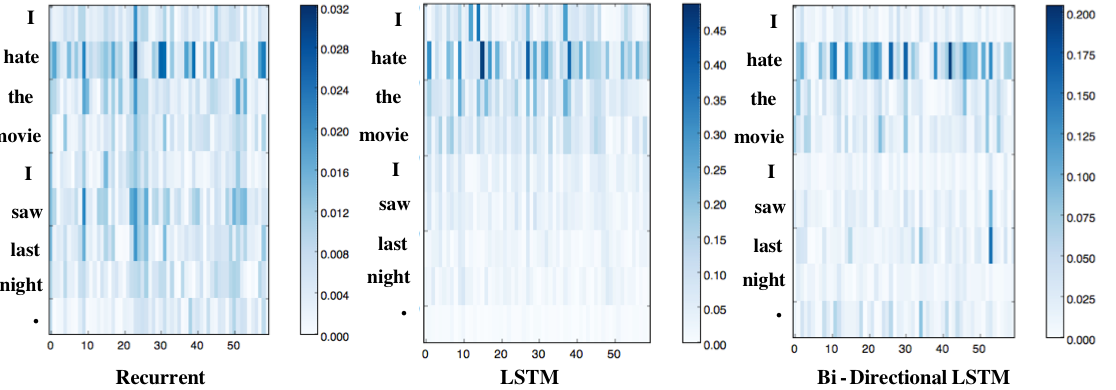}
\caption{Saliency heatmap for ``I hate the movie I saw last night  ." .}\label{V2}
\end{figure*}

\begin{figure*}[!ht]
\centering
\includegraphics[width=6in]{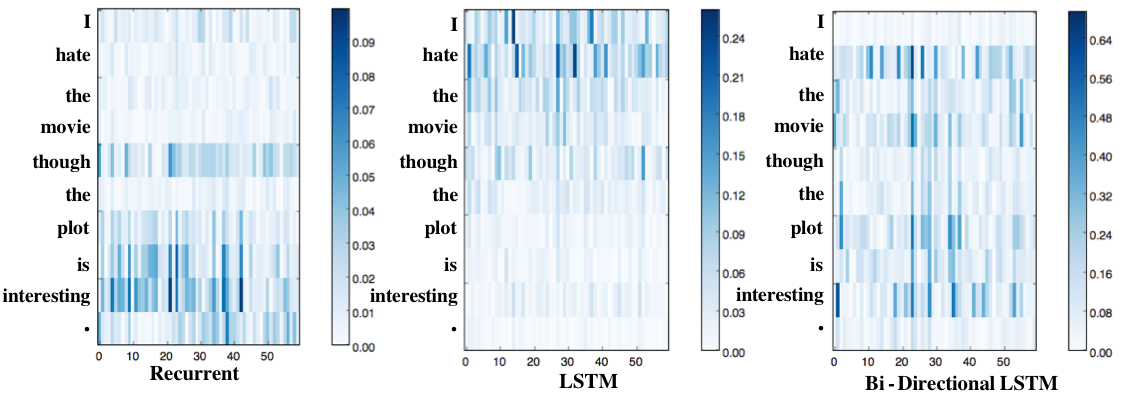}
\caption{Saliency heatmap for ``I hate the movie though the plot is interesting ." .}\label{V3}
\end{figure*}

\section{First-Derivative Saliency}
In this section, we describe another strategy which is 
is inspired by the back-propagation strategy in vision \cite{erhan2009visualizing,simonyan2013deep}.
It measures how much each input unit contributes to the final decision, which can be approximated by 
first derivatives. 

More formally, for a classification model, an input $E$ is associated
with a gold-standard class label $c$. (Depending on the NLP task, an
input could be the embedding for a word or a sequence of words, while
labels could be POS tags, sentiment labels, 
the next word index to predict
etc.)
Given embeddings $E$ for input words with the associated gold class label
$c$, the trained model associates the pair ($E,c$) with a score
$S_c(E)$.  The goal is to decide which units of $E$ make the most
significant contribution to $S_c(e)$, and thus the decision, the choice of class label $c$.

In the case of deep neural models, the class score $S_c(e)$ is a highly non-linear function.
We approximate $S_c(e)$ with a linear function of $e$ by computing the first-order Taylor expansion
\begin{equation}
S_c(e)\approx w(e)^T e+b
\end{equation}
where $w(e)$ is the derivative of $S_c$ with respect to the embedding $e$.
\begin{equation}
w(e)=\frac{\partial(S_c)}{\partial e}\mid_e
\end{equation}
The magnitude (absolute value) of the derivative indicates 
the sensitiveness of  the final decision to the change in one particular dimension, 
telling us how much one specific dimension of the word embedding contributes to the final decision.
The saliency score  is given by
\begin{equation}
S(e)=|w(e)|
\end{equation}
\subsection{Results on Stanford Sentiment Treebank}
We first illustrate  results on Stanford Treebank.
We plot in Figures \ref{V1}, \ref{V2} and \ref{V3} the saliency scores 
(the absolute value of the derivative of the loss function with respect to each dimension of all word inputs)
for three sentences, applying the trained model to each sentence.
Each row corresponds to saliency score for the correspondent word representation with each grid representing each dimension. 
The examples are based on the clear sentiment indicator ``hate" that lends them all negative sentiment.

\paragraph{``I hate the movie"}
All three models assign high saliency  to
``hate" and dampen the influence of other tokens. 
LSTM offers a clearer focus on  ``hate" than the
standard recurrent model, but the bi-directional LSTM shows the clearest
focus, attaching almost zero emphasis on words other than
``hate".   This is presumably due to the gates structures in LSTMs
and Bi-LSTMs that controls information flow, making these architectures
better at filtering out less relevant information.

\begin{figure*}[!ht]
\centering
\includegraphics[width=6.0in]{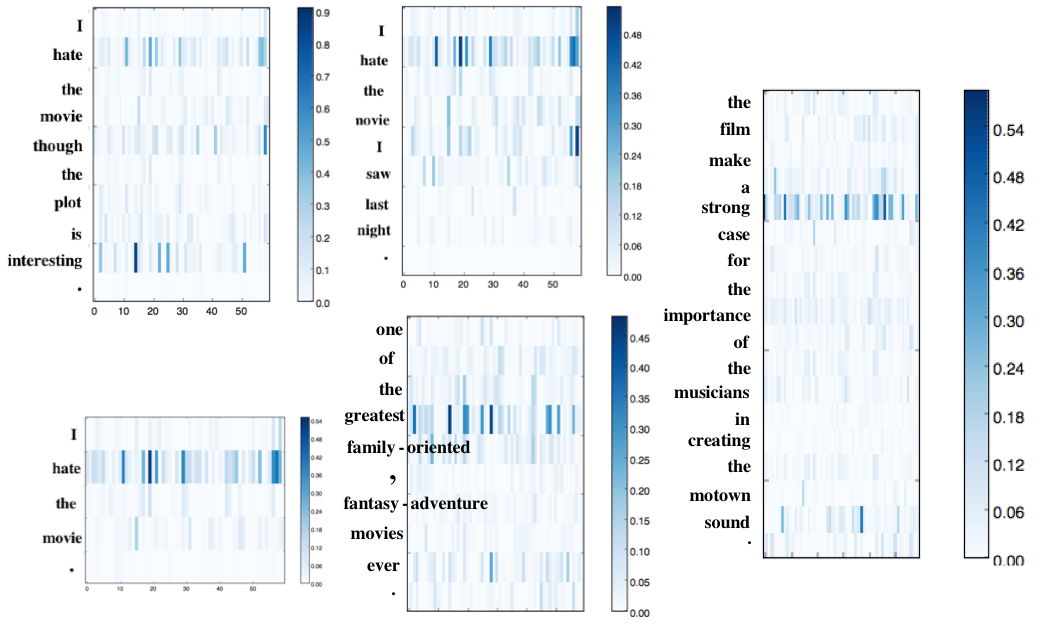}
\caption{Variance visualization. }\label{V4}
\end{figure*}

\paragraph{``I hate the movie that I saw last night"}
All three models assign the correct sentiment.
The simple recurrent models again do poorly at  filtering out irrelevant information,
assigning too much salience to words unrelated to sentiment.
However none of the models suffer from the gradient vanishing problems 
despite this sentence being longer; the salience of
``hate" still stands out after 7-8 following convolutional operations. 

\paragraph{``I hate the movie though the plot is interesting"}
The simple recurrent model emphasizes only the second clause ``the plot
is interesting", assigning no credit to the first clause ``I hate
the movie". This might seem to be caused by a vanishing gradient,
yet the model correctly classifies the sentence as very negative,
suggesting that it is successfully incorporating information from the first negative clause.
We separately tested the individual clause ``though the plot is interesting".
The standard recurrent model confidently labels it as positive.
Thus despite the lower saliency scores for words in the first clause, 
the simple recurrent system manages to rely on that clause and downplay the information from the latter
positive clause---despite the higher saliency scores of the later words.
This illustrates a limitation of saliency visualization.
first-order derivatives don't capture all the information we would like to visualize,
perhaps because they are only a rough approximate to individual contributions 
and might not suffice to deal with highly non-linear cases.
By contrast, the LSTM emphasizes the first clause, sharply dampening the influence from the second clause,
while the Bi-LSTM focuses on both ``hate the movie" and ``plot is interesting". 
\subsection{Results on Sequence-to-Sequence Autoencoder}
\begin{figure}
\centering
\includegraphics[width=3.3in]{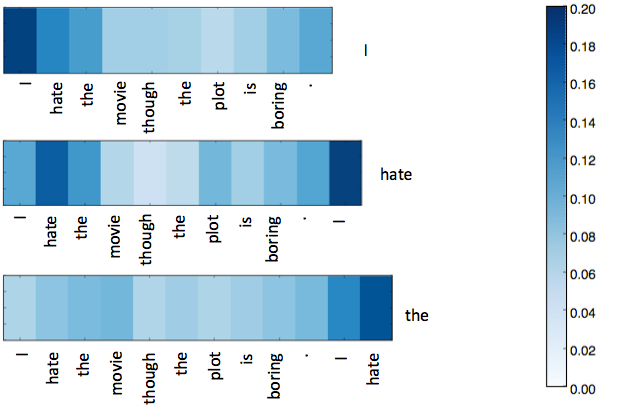}
\includegraphics[width=3.3in]{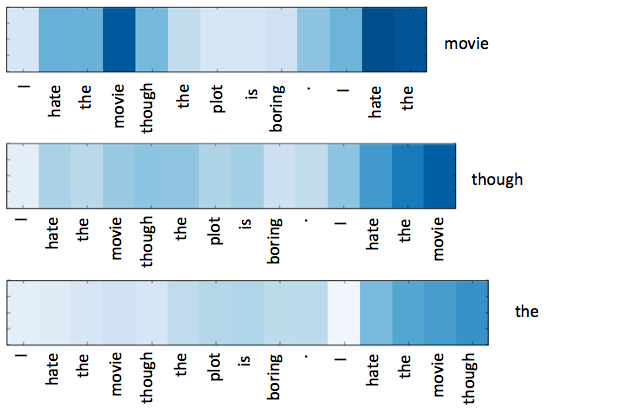}
\includegraphics[width=3.3in]{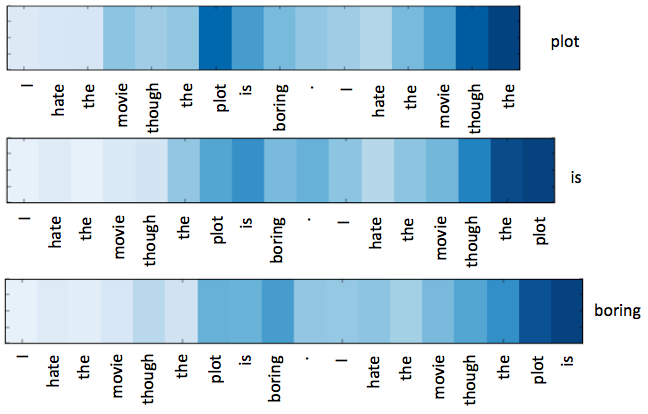}
\caption{Saliency heatmap for \textsc{Seq2Seq} auto-encoder  in terms of predicting correspondent token at each time step.}\label{seq2seq}
\end{figure}

Figure \ref{seq2seq} represents saliency heatmap for 
auto-encoder  in terms of predicting correspondent token at each time step. 
We compute first-derivatives for each preceding word through back-propagation as decoding goes on. 
Each grid corresponds to 
magnitude of 
average saliency value for each 1000-dimensional word vector.
The heatmaps give clear overview about the behavior of neural models during decoding.
 Observations can be summarized as follows:

1. For each time step of word prediction, \textsc{seq2seq} models manage to link word to predict back to correspondent  region at the inputs (automatically learn alignments), e.g., input region centering around token ``hate" exerts more impact when token ``hate" is to be predicted, similar cases with tokens ``movie", ``plot" and ``boring".
 
 2. Neural decoding combines the previously built representation with the word predicted at the current step. 
As decoding proceeds, the influence of the initial input on decoding (i.e., tokens in source sentences) gradually
diminishes as more previously-predicted words are encoded in the vector representations.
Meanwhile,
the influence of language model gradually dominates: when word ``boring" is to be predicted, models attach more weight to 
earlier predicted tokens ``plot"
 and ``is" but less to correspondent regions in the inputs, i.e., the word ``boring" in inputs.   

\section{Average and Variance}
For settings where word embeddings are treated as parameters to
optimize from scratch (as opposed to using pre-trained embeddings),
we propose a second, surprisingly easy and direct way to visualize important
indicators.  We first compute the
average of the word embeddings for all the words within the sentences.
The measure of salience or influence for a word is its deviation from this average.
The idea is that during training, models would learn to render indicators different from
non-indicator words, enabling them to stand out even after many layers of computation.

Figure \ref{V4} shows a map of variance;  each grid
corresponds to the value of $||e_{i,j}-\frac{1}{N_S}\sum_{i'\in N_S}
e_{i'j}||^2$ where $e_{i,j}$ denotes the  value for $j$ th dimension
of word $i$ and N denotes the number of token within the sentences.

As the figure shows, the variance-based salience measure also
does a good job of emphasizing the relevant sentiment words.
The model does have shortcomings: (1) it can only be used in 
to scenarios where word embeddings are parameters to learn 
(2) it's clear how well the model is able to visualize
local compositionality.

\section{Conclusion}
In this paper, we offer several methods to help visualize and interpret neural models, to understand how neural models are able to compose meanings,
demonstrating asymmetries of negation and explain
some aspects of the strong performance of LSTMs at these tasks. 

Though our attempts only touch superficial points in neural models, and each method has its pros and cons, together they may offer some insights into the behaviors of neural models in language based tasks, marking one initial step toward understanding how they achieve meaning composition in natural language processing.
Our future work includes 
using results of the visualization be used to perform error analysis, and understanding strengths limitations of different neural models.

\bibliographystyle{acl}
\bibliography{acl2013}

\noindent {\bf{\large Appendix}}\\
\paragraph{Recurrent Models}
A recurrent network successively takes word $w_t$ at step $t$,
combines its vector representation $e_{t}$ with previously built
hidden vector $h_{t-1}$  from time $t-1$, calculates the resulting
current embedding $h_t$, and passes it to the next step.  The embedding
$h_t$ for the current time $t$ is thus:
\begin{equation}
h_{t}=f(W\cdot h_{t-1}+V\cdot e_{t})
\end{equation}
where $W$ and $V$ denote compositional matrices. 
If $N_s$ denote the length of the sequence, $h_{N_s}$ represents
the whole sequence $S$. 
$h_{N_s}$ is used as input a softmax function for classification tasks. 
\paragraph{Multi-layer Recurrent Models}
Multi-layer recurrent models extend one-layer recurrent structure by operation on a deep neural architecture that enables more expressivity and flexibly. 
The model associates each time step for each layer with a hidden representation $h_{l,t}$, where $l \in [1,L]$ denotes the index of layer and $t$ denote the index of time step. $h_{l,t}$ is given by:
\begin{equation}
h_{t,l}=f(W\cdot h_{t-1,l}+V\cdot h_{t, l-1})
\end{equation}
where $h_{t,0}=e_t$, which is the original word embedding input at current time step. 

\paragraph{Long-short Term Memory}
LSTM model, first proposed in \cite{hochreiter1997long}, maps an input sequence to a fixed-sized vector by sequentially convoluting the current representation with the output representation of the previous step. LSTM associates each time epoch with an input, control and memory gate, and tries to minimize the impact of unrelated information. $i_t$, $f_t$ and $o_t$ denote to gate states at time $t$. $h_{t}$ denotes the hidden vector outputted from LSTM model at time $t$ and $e_t$ denotes the word embedding input at time t. 
We have 
\begin{equation}
\begin{aligned}
&i_t=\sigma(W_i\cdot e_{t}+V_i\cdot h_{t-1})\\
&f_t=\sigma(W_f\cdot e_{t}+V_f\cdot h_{t-1})\\
&o_t=\sigma(W_o\cdot e_{t}+V_o\cdot h_{t-1})\\
&l_t=\text{tanh}(W_l\cdot e_{t}+V_l\cdot h_{t-1})\\
&c_t=f_t\cdot c_{t-1}+i_t\times l_t\\
&h_{t}=o_t\cdot m_t
\end{aligned}
\end{equation}
where $\sigma$ denotes the sigmoid function. $i_t$, $f_t$ and $o_t$ are scalars within the range of [0,1]. $\times$ denotes pairwise dot. 

A multi-layer LSTM models works in the same way as multi-layer recurrent models by enable multi-layer's compositions.  

\paragraph{Bidirectional Models} \cite{schuster1997bidirectional}  add 
bidirectionality to the recurrent framework where embeddings for each time are calculated both forwardly and backwardly:
\begin{equation}
\begin{aligned}
&h_{t}^{\rightarrow}=f(W^{\rightarrow}\cdot h_{t-1}^{\rightarrow}+V^{\rightarrow}\cdot e_{t}) \\
&h_{t}^{\leftarrow}=f(W^{\leftarrow}\cdot h_{t+1}^{\leftarrow}+V^{\leftarrow}\cdot e_{t})
\end{aligned}
\end{equation}
Normally, bidirectional models feed the concatenation vector calculated from both directions $[e_1^{\leftarrow},e_{N_S}^{\rightarrow}]$ to the classifier. 
Bidirectional models can be similarly extended to both multi-layer neural model and LSTM version.

\end{document}